\documentclass[10pt]{wlscirep}

\usepackage{makecell} 
\usepackage{multirow}
\usepackage{colortbl}

\title{Adversarial Training for Disease Prediction from Electronic Health Records with Missing Data}

\author[1]{Uiwon Hwang}
\author[2]{Sungwoon Choi}
\author[3]{Han-Byoel Lee}
\author[1,*]{Sungroh Yoon}
\affil[1]{Department of Electrical \& Computer Engineering, Seoul National University, Seoul, Korea}
\affil[2]{Artificial Intelligence Team, Samsung Electronics Co. Ltd., Seoul, South Korea}
\affil[3]{Department of Surgery, Seoul National University College of Medicine, Seoul, Korea}

\affil[*]{sryoon@snu.ac.kr}

\begin{abstract}
Electronic health records (EHRs) have contributed to the computerization of patient records and can thus be used not only for efficient and systematic medical services, but also for research on biomedical data science. However, there are many missing values in EHRs when provided in matrix form, which is an important issue in many biomedical EHR applications. In this paper, we propose a two-stage framework that includes missing data imputation and disease prediction to address the missing data problem in EHRs. We compared the disease prediction performance of generative adversarial networks (GANs) and conventional learning algorithms in combination with missing data prediction methods. As a result, we obtained a level of accuracy of 0.9777, sensitivity of 0.9521, specificity of 0.9925, area under the receiver operating characteristic curve (AUC-ROC) of 0.9889, and F-score of 0.9688 with a stacked autoencoder as the missing data prediction method and an auxiliary classifier GAN (AC-GAN) as the disease prediction method. The comparison results show that a combination of a stacked autoencoder and an AC-GAN significantly outperforms other existing approaches. Our results suggest that the proposed framework is more robust for disease prediction from EHRs with missing data.
\end{abstract}
\begin{document}

\flushbottom
\maketitle
% ^ <shinyflight@gmail.com> 2018-04-22T06:44:39.738Z.
\thispagestyle{empty}

%\noindent Please note: Abbreviations should be introduced at the first mention in the main text – no abbreviations lists. Suggested structure of main text (not enforced) is provided below.

\section*{Introduction}

% 의료 시스템이 전산화되면서 전자건강기록 (Electric Health Record) [1]은 이전의 서면 의무기록 시스템에 비해 효율적이고 체계적인 의료 서비스가 가능하도록 하는데 크게 기여하였다. 전자건강기록의 중요한 이점 중 하나는 전자건강기록으로부터 생산되는 빅데이터를 통해 질병의 통계학적 분석, 개인화된 질병 예측 등의 다양한 데이터 사이언스 연구를 수행할 수 있다는 점이다.
As medical systems become increasingly computerized, electronic health records (EHRs) \cite{1} are contributing greatly to more efficient and systematic medical services compared to previous written medical record systems. One of the important benefits of EHRs is that big data produced from such records can be used for various data science and machine learning studies \cite{ehr_deep_1, deep1, deep2}, including a statistical analysis of diseases \cite{ehr_stat}, personalized disease prediction \cite{personalized}, and cohort-based disease analysis \cite{ehr_deep_2}.

\begin{figure} [t]
\centering
\includegraphics[width=0.8\textwidth]{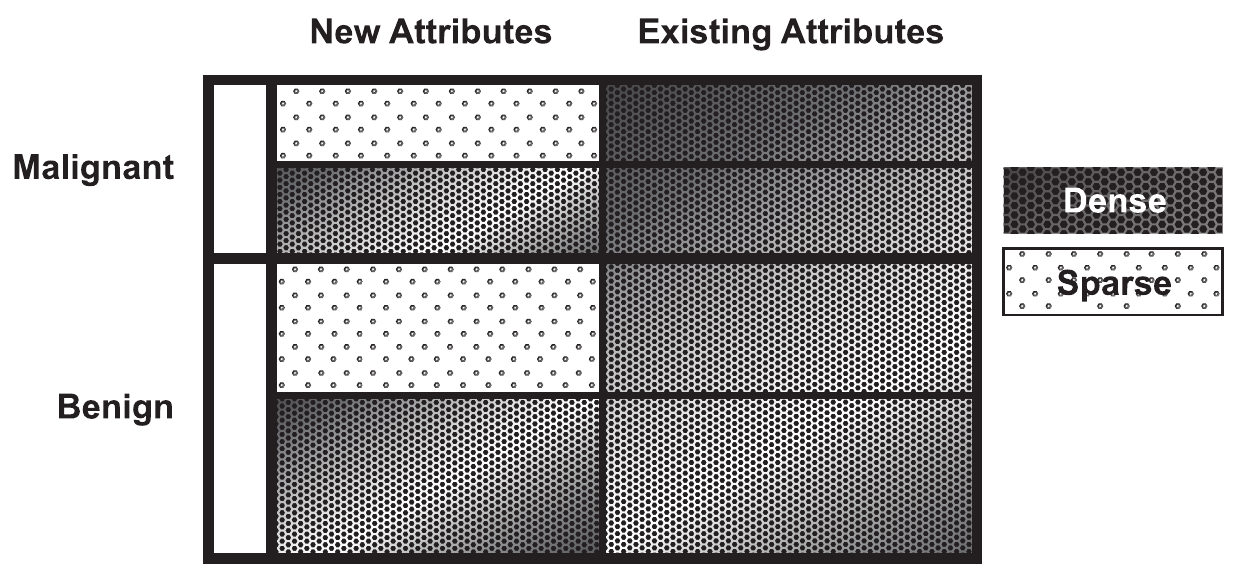}
\caption{\label{fig:sparsity} \textbf{Sparseness in electronic health record data.} This figure shows a case in which new attributes are added to the health examination items and missing values occur in certain attributes of some records.}
\end{figure}

% 전자건강기록을 이용한 데이터 분석에서 발생할 수 있는 문제점은 다음과 같다. 첫째로, 분석할 데이터에 많은 결측치 (Missing data)가 존재하게 된다. 오랫동안 환자들의 건강 상태를 기록하다 보면 의학의 발달 또는 제도의 변경 등으로 인해 특정 시점을 기점으로 새로운 항목에 대한 검사를 시작하는 경우가 많다. 이러한 이유로 전자건강기록의 결측치는 데이터를 수집하는 중에 새로 생겨난 특정 항목에 편중되는 경향이 있다 [2]. 그림 1은 전자건강기록 데이터의 특징적인 희박성 (Sparsity)을 알아보기 쉽게 그림으로 나타낸 것이다.
The problems that can occur in data analysis using EHRs are as follows. First, numerous data in the dataset to be analyzed may be missing. When patients’ health status is recorded over a long period of time, checking on a new attribute often begins at a specific point in time owing to the development of new medicines or changes in the system. For this reason, the missing values in the EHRs tend to be biased toward specific new attributes that were introduced during the data collection process \cite{2}. In addition, there may be many missing values in medical examination data because patients may not have had all of their examinations due to individual needs or cost \cite{missing}. Figure \ref{fig:sparsity} shows a graphical representation of the characteristic sparsity inherent in EHR data. In recent years, platforms have been developed for extracting and organizing data automatically from EHRs \cite{ehr_platform}. In a real clinical setting, however, most data in EHRs are still manually entered into the database, resulting in incorrect or missing values. Thus, EHRs must be reviewed after a list of target patients is made available. The majority of machine learning algorithms do not utilize records that contain missing values. In machine learning studies, it is therefore necessary to impute any missing values before using the data. In previous research, missing values have been filled in with zeros \cite{missing_0} or the average values of each attribute \cite{missing_mean}. A recent study suggested that a deeply learned autoencoder shows a high-level performance for missing data imputation in EHRs \cite{missing_ae}.

% 둘째로는 특정 질병에 걸린 사람의 전자건강기록보다 정상인의 전자건강기록이 일반적으로 더 많으므로 클래스 불균형 문제가 존재한다. 이러한 데이터를 학습에 이용할 때 각 클래스의 샘플 수를 고려하여 손실 함수에 가중치를 주는 기법이나 SMOTE (Synthetic Minority Over-sampling Technique) [3], ADASYN (Adaptive Synthetic Sampling Approach for Imbalanced Learning) [4]과 같은 오버샘플링 (Oversampling) 기법 등으로 어느 정도 완화가 가능하다. 하지만 과적합 (Overfitting)이 일어나거나 학습에 필요한 메모리와 학습 시간을 증가시키는 등의 한계점이 있다 [5]. 따라서 오버샘플링 기법을 이용하지 않고도 클래스 불균형 문제에 더 강인한 학습 모델에 대한 연구가 필요하다.
Another problem that can occur in a data analysis using EHRs is a class imbalance because the number of EHRs for healthy people is greater than that for those with a specific disease. When using such data for learning, we can add class weights to the loss function considering the number of samples of each class \cite{costsensitive}. This can also be mitigated to a certain extent using oversampling techniques such as the synthetic minority over-sampling technique (SMOTE) \cite{3} or adaptive synthetic sampling approach for imbalanced learning (ADASYN) \cite{4}. However, there are certain limitations to these approaches, such as an overfitting or increased memory and time requirements for learning, as summarized by Elrahman et al \cite{5}. It is therefore necessary to develop a learning method that is more robust against the class imbalance problem without using an oversampling technique.

% 생성적 대립 신경망 [6]은 진짜 데이터와 가짜 데이터를 잘 판별하도록 학습되는 판별자 (Discriminator, D)와 판별자를 속일 수 있는 가짜 데이터를 생성하기 위해 학습되는 생성자 (Generator, G)가 서로 경쟁하며 학습되는 알고리즘이다. 생성적 대립 신경망을 이용하여 실제 데이터와 비슷한 통계적 특성과 분류 성능을 갖는 전자건강기록을 생성하는 연구 [7]는 있었지만, 전자건강기록 데이터를 이용한 질병 예측 문제에 생성적 대립 신경망을 이용하는 연구는 현재까지 이루어지지 않았다.
Generative adversarial networks (GANs) \cite{6} are a class of generative models that learn through a competitive process composed of two networks: A discriminator that learns to discriminate between real and fake data, and a generator that learns to generate fake data that can fool the discriminator. Studies on generating EHRs using statistical characteristics and achieving a classification performance similar to actual data using generative neural networks have been conducted \cite{7, 15}. However, to the best of our knowledge, there have been no disease prediction frameworks from EHRs with the assistance of adversarial training and a deep learning based missing data imputation model.

% 본 논문에서는 생성적 대립 신경망을 전자건강기록 데이터에 적용하여 기존에 의료 데이터에 많이 사용되는 서포트 벡터 머신 (SVM)과 질병 예측 성능을 비교하였다. 또한 딥러닝 모델 중 하나인 오토인코더로 전자건강기록 데이터의 결측치를 채워 질병 예측에 이용함으로써 각 항목의 평균값으로 결측치를 채운 경우와 질병 예측 성능을 비교하였다.
In this paper, we suggest a two-stage framework for disease prediction from EHRs with missing values. We compare the predictive performance of auxiliary classifier GANs (AC-GANs) with existing models that are widely used in studies with medical data, such as disease prediction. In the proposed framework, a stacked autoencoder, a type of unsupervised learning algorithm, is used to impute missing data in the EHRs. We also compare the predictive performance of different methods regarding the imputation of missing values.

\section*{Materials and Methods}

This section describes the suggested disease prediction framework, as well as methods used for missing data and disease prediction, the dataset and preprocessing applied, and their implementation.

\subsection*{Disease Prediction from EHRs with Missing Data}
\begin{figure} [t]
\centering
\includegraphics[width=1.0\textwidth]{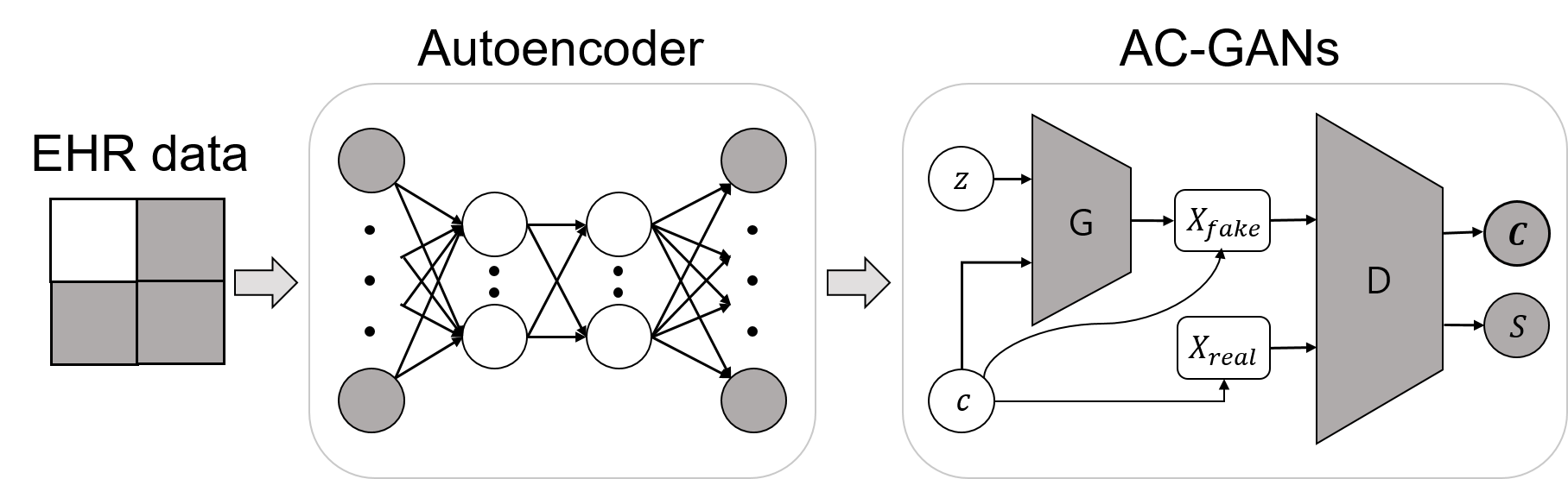}
\caption{\label{fig:overview} \textbf{Diagram of suggested disease prediction framework for EHRs}}
\end{figure}

% 그림 3는 본 논문에서 제안하는 전자건강기록으로부터의 질병 예측 과정을 모식도로 나타낸 것이다. 전체적인 과정을 간략하게 설명하면 다음과 같다. 우리는 전처리된 raw data를 sparsity가 0.1 이상인 records와 그 이하인 records로 나누었다.그리고 sparsity가 0.1 미만인 records를 다시 training data와 validation data로 나누어 autoencoder 학습에 이용하였다. Training data로 autoencoder를 학습하면서 autoencoder의 파라미터에 학습 샘플들의 패턴이 저장되도록 하였다. 검증 샘플을 통해 검증 오류가 최소화되는 Epoch을 찾았다. 그리고 AC-GANs의 학습에 결측치가 존재하는 샘플들을 이용하기 위하여 모든 샘플들을 학습된 모델에 입력시켜서 저장된 패턴 중 가장 적합한 값으로 결측치를 예측하여 채우도록 하였다. AC-GAN은 결측치가 채워진 training data에 의해 환자인지 정상인인지를 예측하도록 학습된다. 학습된 autoencoder와 AC-GAN은 환자의 특정 질병을 효과적으로 예측하게 된다.
Figure \ref{fig:overview} shows a schematic diagram of the disease prediction framework proposed in this paper. A brief description of the overall process is as follows: We split the raw data into records with sparsity of greater than or equal to a specific threshold (e.g., 0.1) and records below this threshold. Records with sparsity of less than the threshold are divided into training and validation data. We train an autoencoder with training data, and patterns of training data are stored in the autoencoder's parameters. Validation data are used to find an epoch that minimizes the validation error. To use samples with missing data in the training of an AC-GAN, missing values of EHR data are filled with the most suitable values using a trained autoencoder. The AC-GAN model is learned to predict whether a record is a patient or a normal person using data whose missing values are imputed by the autoencoder. The trained autoencoder and AC-GAN can efficiently predict the specific disease of the patient.

\subsection*{Missing Data Prediction}
% 결측치를 채우기 위해 두 가지 기법을 이용하였다. 첫 번째는 간단하게 평균치로 채우는 방법이다. 결측치가 있는 데이터를 그대로 질병 예측 알고리즘에 입력하면 학습 시에 오류가 발생하게 되고, 모두 0으로 채우면 이 값들이 결정경계 (Decision boundary)를 정할 때 큰 영향을 주게 된다. 특히 특정 클래스에 결측치가 편중된 경우 질병 예측 모델이 결측치를 채워준 값을 특정 클래스의 패턴으로 학습하여 예측 성능이 떨어지는 결과를 초래하기도 한다. 따라서 결측치에 의한 영향을 최소화하기 위해서 각 항목 (attribute)의 평균치로 결측치를 채워줄 수 있다.

We used Two imputation methods to fill in the missing values. The first method is simply to replace missing values with the mean values of the attributes. When inputting a dataset into a disease prediction algorithm as is, errors occur if the training dataset contains missing values. When all missing values are replaced by zeros, the zeros have a large influence in the decision boundary. When missing values are concentrated in a certain class, it results in a poor prediction performance. Therefore, it is possible to fill in the missing values with the mean values of the attributes to minimize the influence on the decision boundary.

% 두 번째는 비지도 학습 (Unsupervised Learning) 기법의 하나인 오토인코더 (Autoencoder, AE) [8]를 이용하는 기법이다. 오토인코더는 입력값과 같은 출력값을 내도록 식 (1)과 같은 손실함수를 최소화하는 방향으로 모델 파라미터를 학습한다. stacked autoencoder는 히든 레이어가 쌓인 딥 뉴럴 네트워크 구조를 가지고 있다. 이것은 비선형 활성화 함수를 이용하여 데이터 내부의 복잡한 패턴을 계층적인 특징 표현을 통해 잘 학습할 수 있다. 오토인코더의 이러한 특성을 활용하여 기존 결측치 예측 기법들에 비해 더 좋은 성능을 기대할 수 있다. 더 정교한 값으로 결측치가 채워진 데이터는 질병 예측 모델의 학습 및 질병 예측에 이용되어 더 높은 질병 예측 성능을 얻는 데 기여할 수 있다.실험에서 결측치가 거의 없는 샘플들을 학습 샘플과 검증 샘플로 나누어 오토인코더를 학습시킴으로써 모델 파라미터에 학습 샘플들의 패턴이 저장되도록 하고, 검증 샘플을 통해 검증 오류 (Validation error)가 최소화되는 세대 (Epoch)를 찾았다. 그리고 결측치가 있는 샘플들을 학습된 모델에 입력시켜서 저장된 패턴 중 가장 적합한 값으로 예측하여 채우도록 하였다. 오토인코더는 총 5개의 층을 이용하였으며 각 층의 노드 수는 30, 20, 10, 20 30으로 하였다. 또한, 손실 함수로는 이진 교차 엔트로피 (Binary cross entropy)를, 최적화를 위해서는 Adam Optimizer [9]를 이용하였다.

The second method is using an autoencoder (AE) \cite{8}, which is an unsupervised learning algorithm. An autoencoder learns the model parameters to minimize the loss function defined as equation (\ref{eq:aeloss}): 
\begin{equation} \label{eq:aeloss}
L=|| \mathbf{x}_\mathrm{out}-\mathbf{x}_\mathrm{in}||^2,
\end{equation}
where $\mathbf{x}_\mathrm{out}$ is the output of the autoencoder and $\mathbf{x}_\mathrm{in}$ is the input. This is called a reconstruction error, and makes the input and output values of the model equal. A stacked autoencoder has a deep neural network architecture with hidden layers. It can learn complex patterns in data through a hierarchical feature representation using nonlinear activation functions. Using this characteristic of a stacked autoencoder, we can expect a better missing data prediction performance compared with existing missing data prediction methods. Additionally, EHR data with missing values filled in with more suitable values can be used in the training and prediction of disease prediction models, and can contribute to a higher disease prediction performance. In the experiment, samples with few missing values (which means that the missing value ratio of such samples are under the threshold value) were drawn and divided into training and validation datasets. Patterns are stored in the model parameters of an autoencoder by encoding and decoding the training data. The validation set is used to find the epoch where the validation error is minimized. Then, the missing values are imputed with the most appropriate values among the stored patterns using the learned autoencoder. In the proposed framework, we adopt a stacked autoencoder with three hidden layers to cope with missing values.

\subsection*{Disease Prediction}

% 몇가지 방법들이 질병 예측의 비교실험을 위해 사용하였다. Decision Tree는 non-parametric supervised learning 알고리즘으로 data attribute로부터 decision rule을 학습하여 분류를 수행한다. naive bayes classifier는 attribute 간의 독립을 가정하는 bayes 정리를 적용한 알고리즘이다. naive bayes는 의료진단 task에서 적절한 전처리를 통해 SVM과도 경쟁력이 있음을 보여주었다.[ibm] 서포트 벡터 머신 (Support Vector Machine, SVM)은 널리 쓰이는 기계학습 방법의 하나로, 특징 공간 (Feature space) 에서 두 클래스 간의 마진 (Margin)이 최대화되는 초평면을 찾는 알고리즘이다. 서포트 벡터 머신은 커널 함수를 적용하여 특징 공간을 변형함으로써 비선형 분류가 가능하도록 할 수 있는데, 실험에서는 커널 함수로 가우시안 방사 기저함수 (Gaussian Radial Basis Function)를 사용하였다.
Several methods are used for comparative experiments of disease prediction. A decision tree \cite{decisiontree} is a non-parametric supervised learning algorithm that learns decision rules from data attributes and performs classification. A na\"ive Bayes classifier \cite{naivebayes} is a machine learning algorithm that applies Bayes' theorem under an assumed independence between attributes. Na\"ive Bayes shows a similar performance as a support vector machine (SVM) through a proper preprocessing in a disease diagnosis task \cite{nb_ehr}. An SVM \cite{svm} is one of the most widely used machine learning methods. It finds a hyperplane that maximizes the margin between two classes in the feature space. A kernel function \cite{svm_kernel} can be applied to make the SVM separate the data in a nonlinear feature space. In our experiment, we used a Gaussian radial basis function as the kernel function.

% 앙상블 방법들이 여러개의 weak classifier를 학습시킴으로써 일반화 성능을 향상시키기 때문에 Kaggle과 같은 경쟁에서 널리 사용된다. 그중에, Random Forest, AdaBoost, Gradient Boosting이 비교를 위해 사용되었다. Random Forest는 여러 개의 Decision Tree를 학습하는 알고리즘이다. 각 Tree는 전체 attribute 중 일부 attribute만 랜덤으로 이용한다. 실험에서는 각 Tree의 학습에 이용하는 data는 중복을 허용하여 원래 dataset size 만큼 랜덤하게 sampling 하였고, 각 Tree가 5개의 attribute를 랜덤으로 선택하여 학습하도록 하였다. AdaBoost는 Adaptive Boosting이라고도 불리며, 다른 ensemble 모델과 마찬가지로 여러 weak classifier를 학습시킨 뒤 weighted sum하여 final output을 낸다. 그런데 Adaboost는 더 좋은 성능을 위하여 classifier 간의 weight를 adjust하는 방법을 추가한 알고리즘이다. 실험에서는 weak classifier로서 decision tree를 사용하였다. Gradient Boosting은 function space에서 gradient descent를 이용하는 알고리즘이다. original weak classifier의 function space에서의 gradient를 계산한 후, 새로운 모델은 gradient를 target으로 하여 이전 모델의 weakness를 보완하는 역할을 한다.
Ensemble methods \cite{ensemble} have been used extensively in competitions such as a Kaggle' competitions (\href{http://www.kaggle.com/}{http://www.kaggle.com/}) because they improve the generalization power by training multiple weak estimators. In such methods, random forest \cite{randomforest}, AdaBoost \cite{adaboost} and gradient boosting \cite{gradboost} with 10 estimators are used for comparison. Random forest is an algorithm that learns several decision trees as weak estimators. A "tree correlation" \cite{treecorrelation} problem occurs where every tree yields similar outputs when the outputs are determined by certain particular attributes. Therefore, to avoid this problem, each tree uses only some of the attributes. In our experiment, each tree used five randomly selected attributes. Training records were randomly drawn with replacement. AdaBoost, also known as Adaptive boosting, learns several weak classifiers, similar to other ensemble methods. AdaBoost calculates the final output by combining the outputs of weak estimators into a weighted sum, and adjusts the weights between estimators for a better performance. In the experiment, we used decision trees as weak estimators of AdaBoost. Gradient boosting is an ensemble method that uses a gradient descent in the function space. After calculating the gradient in the function space of the current weak estimator, the next estimator compensates for the weakness of the current estimator by setting the gradient as the target. 

% 또한 기본적인 인공신경망으로서 또한 기본적인 인공신경망으로서 MLP가 사용되었다. 비교의 공정성을 위해 AC-GAN의 D와 같은 구조의 MLP를 사용하였다.
In addition, as a basic neural network, a multilayer perceptron (MLP) \cite{mlp} was used. To guarantee the fairness of the comparison, MLP used in the experiment has the same architecture as the discriminator of the AC-GAN.

% 생성적 대립 신경망은 최근 각광받는 딥러닝 모델이며 생성적 모델의 일종이다. 생성자는 노이즈 분포에서 샘플링한 벡터를 입력으로 받아 가짜 데이터를 생성한다. 판별자는 진짜 데이터와 생성자가 생성한 가짜 데이터를 입력으로 받아 입력된 데이터가 진짜인지 가짜인지를 판단하는 역할을 수행한다. 생성자와 판별자는 최소극대화 (minimax) 게임의 내쉬 균형 (Nash equilibrium)을 찾는 방식으로 학습을 수행하며 이를 위해 계산되는 오류 신호 (Error signal)는 식 (2), (3)과 같다. [식 2,3], where
GANs have been popular deep learning models in recent years \cite{gan1, gan2, gan3}, and are a type of generative model for learning a data distribution. The generator $G$ takes the sampled vector from the noise distribution as input, and generates fake data. The discriminator $D$ receives real and fake data generated by the generator as input, and discriminates whether the input data are real or fake. The generator and discriminator learn by finding the Nash equilibrium \cite{nash} of the minimax game, and the error signal is defined as equations (\ref{eq:gan1}) and (\ref{eq:gan2}):
\begin{gather}
\min_{G} \max_{D} V(D,G)  \label{eq:gan1}\\
V(D,G) =  \mathbb{E}_{\mathbf{x}\sim p_\mathrm{data}(\mathbf{x})}[\log D(\mathbf{x})] + \mathbb{E}_{\mathbf{z}\sim p_{\mathbf{z}}(\mathbf{z})}[\log (1-D(G(\mathbf{z})))], \label{eq:gan2}
\end{gather}
where $p_\mathrm{data}$ and $p_{\mathbf{z}}$ represent the data and noise distributions respectively.

% 최근 생성적 대립 신경망의 다양한 변종 모델들이 연구되고 있는데 본 연구에서는 그 중 AC-GANs (Auxiliary Classifier GANs)를 이용하였다. AC-GANs의 생성자는 진짜 데이터 미니 배치 (Minibatch) (X_real)의 클래스 라벨(c)과 같은 값을 노이즈 z의 미니 배치에 붙여서 입력으로 받은 후 가짜 데이터(X_fake)를 생성한다. 판별자는 진짜 데이터와 생성자가 생성한 데이터를 입력으로 받아서 두 개의 출력값을 내는데, 입력된 데이터의 진위를 판별하는 역할을 하는 S와 양성 (Negative)과 악성 (Positive) 클래스를 구별하는 역할을 하는 C로 구성되어 있다. AC-GANs의 학습을 위한 손실 함수를 수식으로 나타내 보면 식 (1), (2)와 같다 [10]. 
% AC-GANs에서 판별자는 L_C+L_S를 최대화하도록 학습되고 생성자는 L_C-L_S를 최대화하도록 학습된다. 이를 통해 생성자는 각 클래스의 데이터 분포와 유사한 데이터들을 생성하도록 발전하고, 이렇게 생성된 각 클래스의 가짜 데이터가 판별자를 속이면서 판별자는 각 클래스의 데이터를 더 잘 분류하도록 발전하게 된다.
Recently, many different variants of GANs have been proposed. In this study, we used auxiliary classifier GANs (AC-GANs) \cite{10}. The generative model $G$ of AC-GANs receives the same value as the class label $c$ of the real data as a condition in addition to the noise vector, and generates fake data. The discriminative model $D$ receives the real data and the data generated by $G$ as inputs and estimates not only the probability that a sample is real or fake but also the class label distribution. All layers except the output layer are shared by $D$. The loss function for AC-GANs consists of two parts and can be expressed as the following equations:
\begin{gather}
\begin{align}
L_s &=  \mathbb{E}[\log P(S = real \ | \ X_{real})] + \mathbb{E}[\log P(S = fake \ | \ X_{fake})]\\
L_c &=  \mathbb{E}[\log P(C = c \ | \ X_{real})] + \mathbb{E}[\log P(C = c \ | \ X_{fake})],
\end{align}
\end{gather} 
where $X_{real}$ and $X_{fake}$ are real and fake data, $S$ denotes whether a sample comes from real data or fake data, and $C$ denotes the class label. In addition, $L_s$ is the same as the original GAN loss function, and $L_c$ is designed to maximize the log-likelihood of the correct class. Finally, $D$ is learned to maximize $L_C+L_S$, and $G$ is learned to maximize $L_C-L_S$. In this way, $G$ is trained to generate data similar to the real data of each class, and $D$ is trained to better classify the data in each class as well as to distinguish the real and fake data.

\subsection*{Dataset}
% 실험 데이터로는 위스콘신 유방암 데이터셋 (Breast Cancer Wisconsin (Diagnostic) Data Set) [11]을 이용하였다. 클래스 라벨은 유방암 양성 (Benign) 357명과 악성 (Malignant) 212명으로 구성되어 있어 클래스 불균형 문제가 일부 존재한다. 원래 데이터셋에서는 0으로 채워진 결측치가 전체 데이터의 약 0.46% 정도밖에 존재하지 않지만, 서론에서 설명한 전자건강기록 데이터의 특징적인 결측치 문제를 발생시키기 위하여 총 30가지 속성 중 절반인 15가지 속성을 임의로 선택하여 각 클래스의 절반의 샘플에서 이 속성값들을 제거하였다.
We used the Breast Cancer Wisconsin (Diagnostic) dataset \cite{11}. The dataset consists of a total of 569 records, with 357 benign and 212 malignant data for breast cancer, which means the dataset has some class imbalance problems. The dataset contains information regarding the characteristics of a patient's cell nuclei obtained using a fine needle aspirate (FNA) of a breast mass. The attributes of the dataset are the mean, standard error, and maximum value of ten features, resulting in a total of 30 attributes. The ten features are the radius (mean of distances from the center to points on the perimeter), texture (standard deviation of gray-scale values), perimeter, area, smoothness (local variation in radius lengths), compactness (perimeter$^2$/area - 1.0), concavity (severity of concave portions of the contour), concave points (number of concave portions of the contour), symmetry, and fractal dimension (coastline approximation \cite{coastline} - 1). In the original data set, only about 0.46\% of the data are missing (filled in with zeros). However, to realize the characteristic missing data problem of the EHRs mentioned in the introduction, we selected and removed the first 15 attributes arbitrarily, namely, half of the 30 attributes in total, from half of the samples in each class.

\subsection*{Preprocessing}
% original dataset에서 label을 분리하였다. benign을 0 (negative)로, malignant를 1 (positive)로 정의하였다. 그리고 각 클래스의 절반의 sample에서 임의로 선택된 attributes를 제거하였다. image processing을 비롯한 많은 분야에서 attribute를 0에서 1사이의 값으로 normalize 한다. 따라서 우리는 제거된 값들을 제외한 attributes를 다음 식을 이용하여 between zero to one으로 scaling하였다: [식] so that 최대절댓값이 unit size로 scale 되게 하면서 처리된 값이 attribute의 std에 대해 robust 하도록 convert 하였다. 그리고 missing values를 각 feature의 평균으로 채운 뒤 저장하여 AC-GAN의 input으로 이용하고, missing value를 0으로 채워 autoencoder를 위한 input data로 이용하였다.
The original dataset was separated into class labels and data. We defined a class label as 0 (negative) for benign and 1 (positive) for malignant. We then removed arbitrarily selected attributes from half of the samples of each class. In many applications of machine learning including image processing, all input data for the neural network model are normalized to values of between 0 and 1. Therefore, we scaled all data except the removed values to a range between 0 and 1 using equation (\ref{eq:scale}):
\begin{equation} \label{eq:scale}
z_i = \frac{x_i-\min(\mathbf{x})}{\max(\mathbf{x})-\min(\mathbf{x})},
\end{equation}
where $\mathbf{x}$ is a column vector of an attribute and consists of $(x_{1}, ..., x_{n})$ which are the patient's records of the attribute, and $\min(\mathbf{x})$ and $\max(\mathbf{x})$ are the maximum and minimum values of the attribute. This method scales the attributes to an equal range so that the converted values are robust to the standard deviation of the attribute data. The data whose missing values were imputed with the mean values of the attributes were used for learning the AC-GAN, and the data whose missing values were imputed with zeros were used as the input of the autoencoder.

\subsection*{Implementation}
% classifiers의 학습과 성능 측정은 5-fold 교차 검증을 통해 이루어졌으며 전체 데이터를 다섯으로 나눈 후 그 중 하나를 평가에 이용하고 나머지들을 학습에 이용하기를 5회 반복하였다. 그리고 일반적인 성능을 비교하기 위해 같은 실험을 10회 반복하여 평균 성능을 계산하였다. 결측치 예측 기법은 Keras [12]를 이용하여 구현하였으며 질병 예측 기법은 Python 라이브러리 중 Scikit-learn [13]과 Tensorflow [14]를 이용하여 구현하였다.
%아키텍쳐, optimizer and 파라미터
All results were obtained using a 5-fold cross-validation procedure \cite{cv}, where all data were divided into five datasets, one of which was used as a test set, and the rest were used for training.
Results were averaged over 10 trials of 5-fold cross-validation to compare the general performance of the AC-GAN and the other disease prediction methods. 
We implemented the stacked autoencoder in Keras \cite{12}, which is a high-level neural networks API. Because Keras supports early stopping \cite{earlystopping}, the training was stopped automatically before convergence to avoid an overfitting. Our stacked autoencoder had five layers including the input and output layers, and the numbers of neurons for each layer were 30, 20, 10, 20, and 30. We optimized the parameters of the autoencoder using a stochastic gradient descent and backpropagation method. The AC-GAN was implemented in Tensorflow\cite{14}. The generator and discriminator of the AC-GAN had a hidden layer with 50 neurons, and used rectified linear units \cite{relu} (ReLU) as an activation function. The output layer of the generator had 30 neurons, which was the same as the number of attributes. The output layer of the discriminator had two neurons, one to distinguish whether the input is real or fake, and the other to predict the class label. We optimized the parameters of the AC-GAN using an Adam Optimizer \cite{9}. The other disease prediction algorithms were implemented in Scikit-learn \cite{13}. In particular, We implemented a multilayer perceptron with a hidden layer with 50 neurons, which was optimized using the Adam Optimizer for comparison under similar experimental conditions.

\section*{Results and Discussion}
% 우리는 결측치를 평균과 오토인코더로 채운 경우 각각에 대해서 질병 예측 기법 별 성능을 비교하였다. 성능은 정확도, sensitivity, specificity, roc curve의 auc, f-score를 포함하는 quantitative evaluation 기법으로 측정되었다. 각 evaluation metric은 다음 식들과 같이 계산된다: [식]. 또한 T-SNE를 이용한 qualitative evaluation 기법을 이용하여 결측치 예측 기법을 평가하였다. T-SNE 설명. 그리고 각 질병 예측 기법에서 결측치 예측 기법에 따른 성능을 비교하였다. 그리고 AC-GAN이 질병 예측 모델로서 우수성을 가지고 있다는 것을 실험을 통해 보여주었다. 본 논문에서 제안하는 파이프라인이 missing data 문제를 가진 EHR에서 질병 예측에 적용될 수 있음을 보였다. 마지막으로 본 논문이 가진 한계점과 이를 극복하기 위한 future work를 논하였다. 
In this section, we compare the performances of disease prediction methods on the mean-imputed data and autoencoder-imputed data. The performances were measured using quantitative evaluation metrics, including the accuracy, sensitivity, specificity, area under the receiver operating characteristic curve (AUC-ROC), and F-score. We also evaluate the missing data prediction methods and the generation power of the AC-GAN generator through qualitative evaluation methods, including Barnes-Hut t-distributed stochastic neighborhood embedding (t-SNE) \cite{tsne} and the ROC curve. Experiments showed that AC-GAN is excellent as a disease prediction model. We discuss the context and significance of the suggested framework for disease prediction from EHRs with missing data. Finally, we discuss the limitations of our research, as well future work to overcome these limitations.

\subsection*{Evaluation on Mean-imputed Data}

\begin{table}[t]
\centering
\renewcommand{\arraystretch}{1.2}
\begin{tabular}{|p{3cm}|>{\centering\arraybackslash}p{2cm}|>{\centering\arraybackslash}p{2cm}|>{\centering\arraybackslash}p{2cm}|>{\centering\arraybackslash}p{2cm}|>{\centering\arraybackslash}p{2cm}|}
\hline
\rowcolor[HTML]{9B9B9B} \textbf{Classifier} & \textbf{Accuracy} & \textbf{Sensitivity} & \textbf{Specificity} & \textbf{AUC-ROC} & \textbf{F-Score} \\
\hline
Decision Tree & 0.9297 & 0.9340 & 0.9272 & 0.9241 & 0.9083 \\
\hline
Na\"ive Bayes & 0.9455 & 0.9245 & 0.9580 & 0.9776 & 0.9267 \\
\hline
SVM (RBF) & 0.9367 & 0.8396 & \textbf{0.9944} & 0.9872 & 0.9082 \\
\hline
AdaBoost & 0.9411 & 0.9311 & 0.9471 & 0.9822 & 0.9218 \\
\hline
Random Forest & 0.9511 & 0.9198 & 0.9697 & 0.9829 & 0.9334 \\
\hline
MLP & 0.9525 & 0.9198 & 0.9720 & 0.9874 & 0.9409 \\
\hline
Gradient Boosting & 0.9596 & 0.9340 & 0.9748 & 0.9870 & 0.9451 \\
\hline
\textbf{AC-GAN} & \textbf{0.9752} & \textbf{0.9498} & 0.9898 & \textbf{0.9886} & \textbf{0.9647} \\
\hline
\end{tabular}
\caption{\label{tab:Results_mean} \textbf{Performance evaluation of disease prediction methods on mean-imputed data.}}
\end{table}

\begin{table}[t]
\centering
\renewcommand{\arraystretch}{1.2}
\begin{tabular}{|p{3cm}|>{\centering\arraybackslash}p{2cm}|>{\centering\arraybackslash}p{2cm}|>{\centering\arraybackslash}p{2cm}|>{\centering\arraybackslash}p{2cm}|>{\centering\arraybackslash}p{2cm}|}
\hline
\rowcolor[HTML]{9B9B9B} \textbf{Classifier} & \textbf{Accuracy} & \textbf{Sensitivity} & \textbf{Specificity} & \textbf{AUC-ROC} & \textbf{F-Score} \\
\hline%\hline
Decision Tree & 0.9315 & 0.9151 & 0.9412 & 0.9455 & 0.9087 \\
\hline
Na\"ive Bayes & 0.9350 & 0.8868 & 0.9636 & 0.9814 & 0.9104 \\
\hline
SVM (RBF) & 0.9508 & 0.8774 & \textbf{0.9944} & 0.9874 & 0.9273 \\
\hline
AdaBoost & 0.9490 & 0.9292 & 0.9608 & 0.9847 & 0.9314 \\
\hline
Random Forest & 0.9561 & 0.9340 & 0.9692 & 0.9834 & 0.9406 \\
\hline
MLP & 0.9578 & 0.9245 & 0.9776 & 0.9883 & 0.9423 \\
\hline
Gradient Boosting & 0.9613 & 0.9387 & 0.9748 & 0.9883 & 0.9476 \\
\hline
\textbf{AC-GAN} & \textbf{0.9777} & \textbf{0.9521} & 0.9925 & \textbf{0.9889} & \textbf{0.9688} \\
\hline
\end{tabular}
\caption{\label{tab:Results_ae} \textbf{Performance evaluation of disease prediction methods on autoencoder-imputed data.}}
\end{table}

\begin{figure} [p]
\centering
\includegraphics[width=1.0\textwidth]{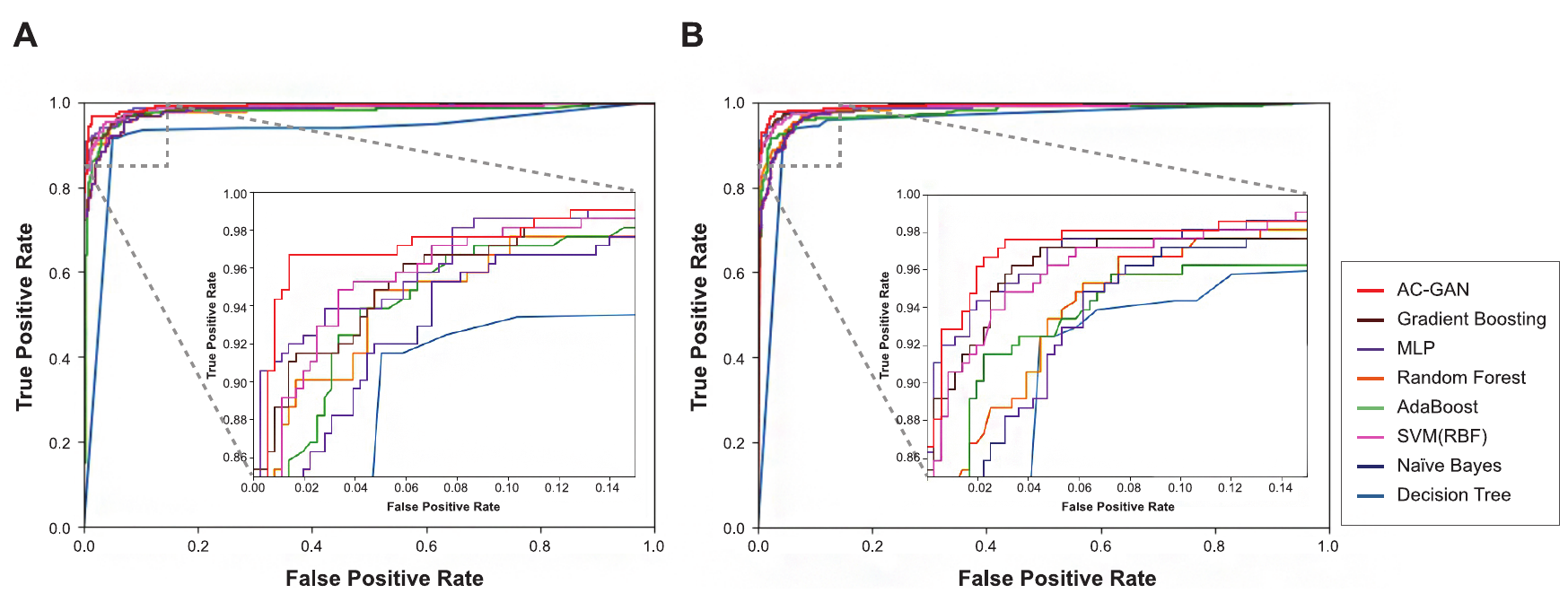}
\caption{\label{fig:ROC}\textbf{Comparison of ROC curves of disease prediction methods.} This figure shows the ROC curves at full scale along with zoomed-in views of upper-left region of the plots. \textbf{(A)} ROC curves obtained using disease prediction methods on mean-imputed data \textbf{(B)} ROC curves obtained using disease prediction methods on autoencoder-imputed data}
\end{figure}

\begin{figure} [p]
\centering
\includegraphics[width=1.0\textwidth]{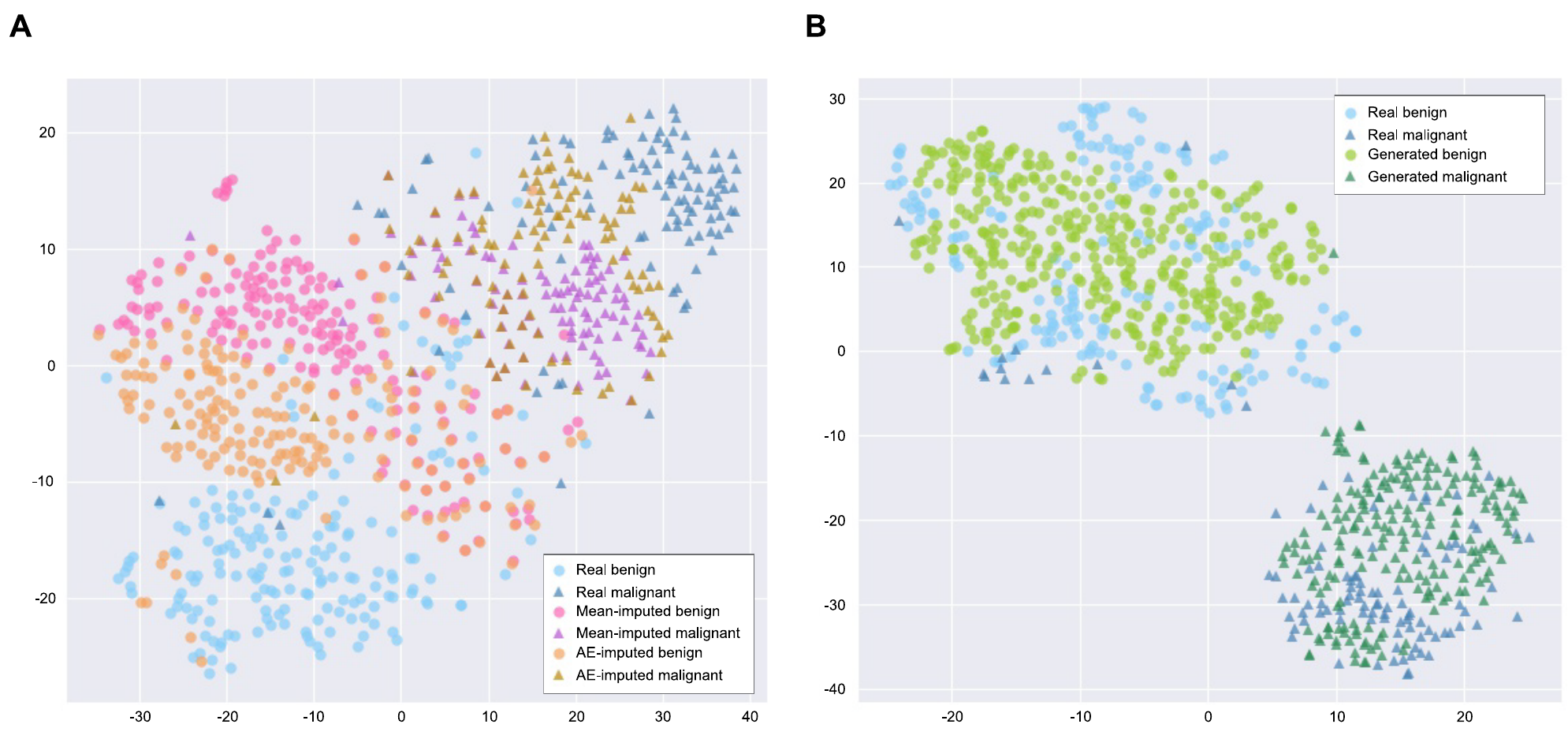}
\caption{\label{fig:TSNE}\textbf{t-SNE map of electronic health records.} It should be noted that a 2D map generated using t-SNE is determined by training a neural network, resulting in different mappings for different initializations. \textbf{(A)} 2D map of the real data (Real benign and Real malignant) and data whose missing values were filled in with the mean value (Mean-imputed benign and Mean-imputed malignant) and using the trained autoencoder (AE-imputed benign and AE-imputed malignant), which shows the performances of the missing data prediction methods qualitatively. \textbf{(B)} 2D map of the real data (Real benign and Real malignant) and data generated using the generator of the trained AC-GAN (Generated benign and Generated malignant).}
\end{figure}

% %\multicolumn{7}{l}{\scriptsize TP: $\sum true \ positive$, TN: $\sum true \ negative$, FP: $\sum false \ positive$, FN: $\sum false \ negative$} \\[-0.5ex]
%\multicolumn{7}{l}{\scriptsize $^{1}$ACC (accuracy) = (TP+TN) / (TP+TN+FP+FN)    $^{2}$SE (sensitivity) = TP / (TP+FN)    $^{3}$SP (specificity) = TN / (TN+FP)} \\[-0.5ex]
% 질병 예측 방법들을 비교하기 위하여 다음과 같은 evaluation metrics를 이용하였다.
%각 metric 설명 : Accuracy는 전체 dataset 중 클래스 라벨을 맞춘 비율로 계산되며 전체 sample 중 true positive와 true negative 수의 비율이다. sensitivity는 실제 환자를 환자로 예측하는 비율을 나타내며 OO 대비 OO의 비율로서 계산된다. specificity는 정상인을 정상인으로 예측하는 비율을 나타내며 OO 대비 OO의 비율로서 계산된다. ROC curve는 true positive rate versus false positive rate의 plot이다. AUC-ROC는 ROC curve를 [0,1]범위에서 적분하여 계산한 것이다. F-score는 precision과 recall의 조화평균으로 다음과 같이 계산된다. [f-score=precision,recall=tf,tp], where tf는 ~, tp는 ~, precision은 p()를 의미하며 recall 은 p()를 의미한다.
The following evaluation metrics were used to evaluate the disease prediction methods: The accuracy is defined as the ratio of correct classifications to the total number of samples examined, and is calculated as the number of true positives and true negatives divided by the number of all results. The sensitivity \cite{sen} of a disease prediction is the ability to correctly identify those patients with a disease, and is calculated as the number of true positives divided by the number of patients who actually have the disease. The specificity \cite{spe} of a disease prediction is the ability to correctly identify those patients without a disease, and is calculated as the number of true negatives divided by the number of patients who actually do not have the disease. An ROC curve \cite{roc} is a plot that illustrates the diagnostic capability of a binary classifier system, considering multiple possible thresholds. AUC-ROC is calculated by integrating the ROC curve. The F-score \cite{fscore} (F$_{1}$ score) is the harmonic average of the precision and recall, where precision is the proportion of true positives among the positive results, and recall is the same as the sensitivity.

% 결과 비교 : 두 가지의 결측치 예측 기법과 두 가지의 질병 예측 (분류) 기법의 조합에 대한 성능 측정 결과는 표 1과 같다. 그리고 두 분류기 모두 민감도에 비해 특이도가 더 높은 경향을 보인다. 이는 양성이 악성에 비해 데이터에 더 많은 클래스 불균형 문제가 있기 때문에 양성 (Negative)으로 예측하는 빈도가 높아서 이러한 결과가 나타난 것으로 예상된다. AC-GANs는 0.9373과 0.9735로 두 질병 예측 모델 간에 큰 차이를 보이는 것은 AC-GANs가 클래스 불균형 문제에 더 강인하여 실험에 이용한 데이터의 소수 클래스인 악성 (Positive)를 잘 예측하였기 때문이라고 해석할 수 있다. 특히 질병을 예측하는 문제에서는 실제 환자를 환자로 예측하는 비율인 민감도를 높이는 것이 중요하기 때문에 AC-GANs이 서포트 벡터 머신에 비해 더 좋은 질병 예측 성능을 지니고 있다고 생각할 수 있다. 두 질병 예측 기법의 특이도를 비교하였을 때는 오히려 서포트 벡터 머신이 평균값으로 결측치를 채웠을 때와 오토인코더를 이용하여 채웠을 때 동일하게 0.9916으로 0.9689와 0.9861의 특이도를 보인 AC-GANs보다 높은 값을 보이고 있음을 확인할 수 있다. 이는 서포트 벡터 머신이 클래스 불균형 문제에 강인하지 못하여 상당수의 평가 데이터를 양성 (Negative)으로 예측하기 때문에 오히려 특이도가 높아지는 현상을 보인다고 생각할 수 있다. AC-GAN이 약간 더 좋은 AUC-ROC 점수를 보였다. AUC-ROC 점수는 모든 threshold를 고려하기 때문에 threshold가 달라질 때 전반적으로 좋은 성능을 보임을 알 수 있다. 그림 3 A의 roc curve를 보면,  model의 모든 threshold에서 AC-GAN이 outperform한다고 볼 수는 없다. 하지만 이러한 영역은 거의 관찰되지 않았다. 특히 F-score는 두번째로 높은 방법과 비교하여 2%p 이상의 차이로 outperform 함을 보여주었다. 
% roc curve A 언급 : AUC-ROC 전후
Table \ref{tab:Results_mean} shows the performance measurements of disease prediction models on the mean-imputed data. AC-GAN shows the highest accuracy, namely, 0.9752, for the disease prediction models. In every model, the results showed a tendency toward higher specificity than sensitivity. This tendency is expected due to the high frequency of negative (benign) predictions because a class imbalance problem exists in which the number of benign samples is more than the number of malignant samples in the dataset. The sensitivity of AC-GAN, whose highest value among the models is 0.9498, indicates that AC-GAN is more robust against class imbalance problems. AC-GAN thus predicts positive (malignant) samples well, which are the minority class of data used in the experiments. In particular, in the case of disease prediction, it is important to increase the sensitivity, because sensitivity is the proportion of patients with the disease who are correctly identified by the model. Therefore, AC-GAN has a better disease predictive capability than the other disease prediction models. Comparing the specificity of the disease prediction models, the SVM shows the highest specificity among the disease prediction models, with a specificity of 0.9944. The largest gap between sensitivity and specificity occurs in the SVM. These results indicate that the SVM is not robust against the class imbalance problem. Thus, a large number of patients are predicted to be negative, which results in higher specificity. AC-GAN reported a slightly better AUC-ROC score than the other models. Because the AUC-ROC score considers all discrimination thresholds, it can be seen that the overall performance of AC-GAN for all thresholds is slightly better. As shown in Figure \ref{fig:ROC} (A), AC-GAN does not outperform the other models as its threshold is varied because the ROC curve of AC-GAN intersects with that of the other models at certain points. However, these points are rarely observed in the ROC curve. In addition, the F-score shows about a 2 percentage point improvement at a threshold of 0.5, which is commonly used.

\subsection*{Evaluation on Autoencoder-imputed Data}
% ac-gan은 spe를 제외한 모든 metric에서 모든 실험을 통틀어 가장 좋은 성능을 보여주었다. 결과의 대부분의 경향은 mean-imputed data에서의 결과와 유사한 양상을 보인다. auc-roc는 MLP 및 Gradient Boosting과 비교하여 아주 작은 향상을 보여주고 있다. roc curve는 ac-gan이 다른 모델보다 좋은 성능이 나타남을 보여주고 있지만, 그 격차가 작다는 것 역시 보여주고 있다. 반면에 acgan의 accuracy 및 f-score는 improve far and away the best performance of 다른 질병 예측 모델. 특히 F-score는 두번째로 높은 방법과 비교하여 2%p 이상의 차이로 outperform 함을 보여주었다. 
The performance measurements of the disease prediction models on the autoencoder-imputed data are reported in Table \ref{tab:Results_ae}. Most tendencies in the results are similar to those on the mean-imputed data. For all experiments, AC-GAN shows the best performance in all evaluation metrics except specificity. The specificity of AC-GAN is slightly lower than that of the SVM. The ROC curves in Figure \ref{fig:ROC} (B) indicate that AC-GAN demonstrates a better performance than the other models, but also show that the performance gap is small. The AUC-ROC score of AC-GAN consequently shows a slight improvement compared to MLP and gradient boosting. On the other hand, the accuracy and F-score of AC-GAN are far and away the best performance of the other disease prediction models. In particular, large improvements in the F-score (i.e., over a 2 percentage point) are achieved compared to the second-highest score.

\subsection*{Comparison of Missing Data Prediction Methods}
% 결측치 예측 기법을 비교해 보면 평균값으로 결측치를 채우는 방법에 비해 오토인코더를 이용하는 방법이 서포트 벡터 머신과 AC-GANs에서 더 높은 정확도와 민감도를 보여주었다. 이는 단순히 결측치가 분류기에 영향을 덜 주도록 평균치로 채운 경우에 비해서 나머지 속성들과 연관된 패턴으로 결측치를 채우는 것이 더 많은 정보를 분류기에 제공하여 분류기의 성능을 높이기 때문이라고 이해할 수 있다. Figure 4 (A)에서 t-sne를 이용한 qualitative analysis를 수행하였다. real data와 mean-imputed data, ae-imputed data를 2D space 상으로 mapping 해보면 benign과 malignant 모두에서 ae-imputed data가 mean-imputed data에 비해 real data에 더 가까이 mapping 됨을 확인할 수 있다.
Comparing the missing data prediction methods, the results of the evaluation metrics show a higher performance on the autoencoder-imputed data than on the mean-imputed data in every model except the na\"ive Bayes classifier. Therefore, it can be understood that filling in missing values using the pattern associated with the remaining attributes, as compared with filling in the missing values using the mean value for a lesser effect on the classifier, provides more information to the classifier. This therefore, enhances the classifier performance. The autoencoder does not lead to any specificity improvements in the SVM, but the improved sensitivity results in an increased performance and robustness to the class imbalance problem. In na\"ive Bayes, with the exception of AUC-ROC, the evaluation metrics show lower performances on the autoencoder-imputed data than on the mean-imputed data. However, the higher AUC-ROC values on the autoencoder-imputed data (0.9776 versus 0.9814) suggest that better results than on the mean-imputed data can be obtained by changing the discrimination threshold. Figure \ref{fig:TSNE} (A) visualizes the qualitative analysis on the missing data prediction methods using t-SNE. The 2D map of the real data, mean-imputed data, and autoencoder-imputed data shows that the autoencoder-imputed data are more closely mapped to the real data than the mean-imputed data for both benign and malignant data.

\subsection*{Excellence of AC-GAN as a Disease Prediction Model}
% GAN의 효용성 (다른 알고리즘보다 좋을 뿐만 아니라 Discriminator가 같은 MLP와 비교, fig4.B)
% 처음에 AC-GANs는 mode collapse에 빠지지 않고 다양한 클래스의 이미지를 잘 생성하는 생성자의 학습에 집중하는 모델로 고안되었다. 그러나 본 논문에서는 AC-GANs의 생성자가 양성과 악성에 해당하는 가짜 데이터들을 골고루 생성하여 판별자의 분류 성능을 향상시키는 점에 집중하여 판별자가 학습되면서 강인한 질병 예측이 가능하도록 AC-GANs를 활용하였다. mean-imputed data에서와 autoencoder-imputed data에서의 결과들 둘다 AC-GAN이 다른 모델들에 비해 성능이 좋다는 것을 보여주었다. 특히 AC-GAN의 discriminator와 같은 구조를 사용한 MLP의 결과를 보면 adversarial training이 질병 예측에서 성능 향상에 기여함을 알 수 있다.
In the original paper, AC-GANs were designed as a model focusing on the learning of a generator that generates images of various classes well without causing a mode collapse. In this paper, however, we focused on improving the classification performance of the discriminator by the generator of the AC-GAN generating both benign and malignant data, which makes robust disease prediction possible. The results on both the mean-imputed data and autoencoder-imputed data demonstrate that the AC-GAN outperforms the other disease prediction models. In particular, compared to a multilayer perceptron, which has the same architecture as the discriminator of the AC-GAN, AC-GAN shows that the adversarial training further boosts the performance of disease prediction. We used t-SNE to visualize a 2D map of real data and generated data to qualitatively ascertain how realistic the generated samples synthesized by the generator are. The map is shown in Figure \ref{fig:TSNE} (B). With the t-SNE map, we observed that the generated samples were almost completely mixed with real samples for both benign and malignant data.

\subsection*{Context and Significance}
% Context: abstract 복붙, 딥러닝은 최근 질병예측, genomics, 약물추천 등의 데이터 사이언스 태스크를 수행하는 데에 활용되고 있다. 그런데 행렬 형태로 주어지는 전자건강기록에는 많은 결측치가 존재하는 문제가 있고, 이는 biomedical의 많은 응용에서 중요한 문제이다. 우리가 아는 한, adversarial training의 도움을 받아 더 정확한 질병 예측을 하면서도 missing value imputation을 수행하는 방법론은 현재까지 없었다.
Electronic health records have contributed to the computerization of patient records, and thus they can be used not only for efficient and systematic medical services, but also for research on biomedical data science. Deep learning has recently been applied to biomedical data science tasks such as disease prediction, genomics \cite{genomics, nature_min}, and drug discovery \cite{drug}. However, there are many missing values in EHRs given in matrix form, which is an important issue in many biomedical applications of EHR data. To the best of our knowledge, there are currently no disease prediction frameworks for conducting a more accurate disease prediction through the assistance of adversarial training and a deep learning based missing data imputation model.

% 본 논문에서는 결측치가 있는 전자건강기록에서 질병 예측을 위한 알고리즘의 성능을 분석하였다. 그 결과 결측치를 단순히 평균값으로 채우는 경우보다 오토인코더를 이용하는 경우의 정확도가 더 높았다. 결측치 예측과 질병 예측 기법들의 조합 중 가장 성능이 좋은 조합은 본 논문에서 제안하는 오토인코더로 결측치를 예측한 후 AC-GANs를 이용하여 질병을 예측하는 방법이며 정확도는 0.9735, 민감도는 0.9460, 특이도는 0.9861로 가장 높은 정확도와 민감도를 보여주었다. 기계학습에서 널리 사용되는 서포트 벡터 머신보다 최근에 생성 및 분류 모델로 각광받는 생성적 대립 신경망이 클래스 불균형 문제에 더 강인하며 의료 데이터에서 더 좋은 질병 예측 성능을 보임을 확인하였다.
In this paper, we analyzed the performance of different algorithms for predicting diseases in EHRs with a missing data problem. As a result, the accuracy of filling in the missing values with a stacked autoencoder is higher than that of simply filling in missing values with the mean value. In addition, under the same conditions without oversampling methods, which consume additional memory, generative adversarial networks outperform other disease prediction methods. The best combination of missing data prediction methods and disease prediction methods is to predict missing values with a stacked autoencoder and to then predict a disease using an AC-GAN, which shows a level of accuracy of 0.9777, sensitivity of 0.9521, specificity of 0.9925, AUC-ROC score of 0.9889, and F-score of 0.9688. The results demonstrate that the proposed framework achieves a better disease prediction performance than existing methods widely used for EHR data. In conclusion, the proposed framework is more robust against class imbalance and a missing data problem than existing methods.

% +미래지향적인 뭔가: 제안된 framework을 이용하면 환자 데이터가 실시간으로 들어올 때 온라인으로 질병을 예측하며 데이터를 학습에 이용할 수 있다. 그리고 generator가 reasonable한 가짜 데이터를 만들 수 있으므로 generator의 파라미터에 환자 데이터베이스를 저장할 수 있는 잠재력이 있다. 이는 실제 ehr과 유사한 특성을 갖는 데이터를 생성하여 데이터사이언스 연구에 활용할 수 있다는 이점이 있다. 
As an application of the proposed disease prediction framework, applying it online will make it possible to combine disease prediction with the learning of the framework. In the future, the generator will be able to create more realistic data using advanced algorithms, and there is a potential to store an EHR database in the parameters of the generator. The generator will be able to generate synthesized data whose characteristics are almost the same as real data, and will be a powerful tool for biomedical data science research.

\subsection*{Limitations and Future Work}
% EHR에 주로 나타나는 두가지 문제인 클래스 결측치 문제와 클래스 불균형 문제에 더 강인해지기 위하여 몇 가지 후속연구를 생각해 볼 수 있다. This work에서는 autoencoder를 따로 학습시켜 결측치 예측에 사용하였으나, GAN이 생성 모델이라는 점을 활용하여 generator 자체를 missing data를 채우는 용도로 사용하거나, generator에 대해 constrained된 모델을 함께 학습하면서 missing data를 채우도록 할 수 있다. 추가로 우리는 실험에서 결측치 문제를 발생시키기 위해 지울 attributes를 arbitrarily하게 선택했다. 데이터의 특성을 더 면밀히 파악함으로써 우리는 현실적으로 missing이 발생할 수 있는 attribute를 선택하여 제거할 수 있을 것이다.
Several follow-up studies can be considered to make generative adversarial networks more robust against missing values and class imbalance problems. There are two main problems in our current framework. First, an AC-GAN for disease prediction, and a stacked autoencoder for missing data prediction, are trained separately. However, using the GAN generator itself as a generative model, or using a model whose parameters are constrained by those of the generator, missing-data and disease prediction tasks can help solve each other in adversarial learning. This will enable end-to-end learning by combining the two stages of the current framework together. We plan to implement this idea as a future work. In addition, we arbitrarily selected the attributes to remove that incur the missing data problem. By more thoroughly examining the attributes of the EHRs, attributes that can cause missing values in real clinical settings can be eliminated.

% 클래스 불균형 문제도 generated data로 해결하는 방법을 생각해 볼 수 있다. ACGAN의 학습을 간단하게 발전시킬 수 있는 방안으로 generator에 주어지는 class condition을 real data의 class ratio와 opposite하게 선택해서 discriminator에 들어가는 batch의 class가 balanced되도록 할 수 있다. 이 방법을 통해 Generator가 전체 data의 distribution을 학습하면서도 각 minibatch별로 oversampling을 수행하므로 추가적인 메모리 소모가 거의 없다. 다만 각 class에 대한 진짜와 가짜 data의 balance가 깨지기 때문에 GAN training이 불안정해 질 수 있어 정교한 learning procedure 수정이 필요할 것이다.
Second, the framework suggested in this paper does not directly address the class imbalance problem. The class imbalance problem can also be solved using data generated from a GAN. As an idea for developing an AC-GAN algorithm, it is possible to modify the class conditions that are given to the generator as opposite the class ratio of a mini-batch of real data so that the classes of the mini-batch entering the discriminator can be balanced. In this way, while the generator is learning the distribution of all data, it applies an oversampling for each mini-batch, thereby enabling an oversampling with little additional memory consumption. However, because the balance between real and fake data for each class is broken, GAN training may become unstable. Thus, a sophisticated modification in the learning procedure may be necessary.

\bibliography{sample}

\section*{Acknowledgements}

This work was supported in part by the Future Flagship Program(10053249, Development of Personalized Healthcare System Exploiting User Life-log and Open Government Data for Business Service Model Proof on Whole Life Cycle Care) funded by the Ministry of Trade, Industry \& Energy(MOTIE, Korea), and in part by the National Research Foundation of Korea (NRF) grant funded by the Korea government (Ministry of Science and ICT) [2018R1A2B3001628].

\section*{Author contributions statement}

U.H. designed and conducted experiments, analyzed results, and wrote the manuscript. S.C. designed experiments and refined the manuscript. H.L. provided medical consultation and refined the manuscript. S.Y. conceived the project, supervised research, and refined the manuscript. All authors reviewed and approved the manuscript.

\section*{Additional information}

\textbf{Competing interests:} The authors declare no competing interests.

The corresponding author is responsible for submitting a \href{http://www.nature.com/srep/policies/index.html#competing}{competing financial interests statement} on behalf of all authors of the paper. This statement must be included in the submitted article file.

\end{document}